\documentclass[a4paper,fleqn]{cas-dc}
\usepackage{float}
\usepackage{graphicx}
\usepackage{adjustbox}
\usepackage[numbers,sort&compress]{natbib}
\usepackage{amsmath}
\usepackage{hyperref}
\usepackage{xcolor}
\usepackage{pgfplots}
\pgfplotsset{compat=1.18}

\usepackage{booktabs}
\usepackage{tikz}
\usetikzlibrary{arrows.meta, positioning, fit, calc, shapes.geometric, backgrounds}
\def\tsc#1{\csdef{#1}{\textsc{\lowercase{#1}}\xspace}}
\tsc{WGM}
\tsc{QE}

\begin{document}
\shorttitle{Multimodal Language Models on the NRC Reactor Operator Licensing Examination}
\shortauthors{I. Hwang et al.}
\title[mode = title]{Multimodal Language Models Benchmarked Against the NRC Reactor Operator Licensing Examination: Fine-Tuning and Retrieval Strategies}
\author[1]{Isak Hwang}
\fnmark[1]
\credit{Conceptualization, Methodology, Software, Investigation, Writing -- original draft}
\affiliation[1]{organization={Department of Nuclear Engineering, Hanyang University},
            addressline={222 Wangsimni-ro},
            postcode={04763},
            state={Seongdong-gu},
            city={Seoul},
            country={South Korea}}
\author[2]{Yoon Pyo Lee}[orcid=0009-0004-4044-4447]
\fnmark[1]
\credit{Conceptualization, Supervision, Software, Writing -- review \& editing}
\affiliation[2]{organization={The Grainger College of Engineering, Nuclear, Plasma \& Radiological Engineering, University of Illinois Urbana-Champaign},
            city={Urbana},
            state={IL},
            country={USA}}
\author[2]{Syed Bahauddin Alam}
\cormark[1]
\ead{alams@illinois.edu}
\credit{Supervision, Writing -- review \& editing}

\cortext[1]{Corresponding author}
\fntext[1]{These authors contributed equally to this work.}
% =========================================================
% ABSTRACT
% =========================================================
\begin{abstract}
Competence claims for a language model in a safety-critical domain are credible when measured against a standard the domain already enforces. We evaluate an open-weight 31-billion-parameter multimodal model (Gemma~4 31B-IT) on the U.S.~Nuclear Regulatory Commission Reactor Operator Generic Fundamentals Examination (GFE), scoring it paper by paper against the 80\% criterion applied to every human candidate, with no rounding up. The evaluation set is a census of every GFE administered at the March sitting from 2015 to 2021, giving seven pressurized water reactor (PWR) and seven boiling water reactor (BWR) papers and 697 scored items. Eight configurations cross three model states, the base model, supervised fine-tuning (SFT) on distilled chain-of-thought rationales and retrieval-augmented fine-tuning (RAFT), with three retrieval conditions, none and BM25 retrieval over the Department of Energy Fundamentals Handbooks under fixed-size and structure-aware chunking. Out of the box it answers 51.94\% correctly and passes no paper. SFT with fixed-size chunking retrieval passes 8 of 14, reaching 80.23\% on PWR items and 79.77\% pooled, with a Wilson interval spanning the threshold. The preferred chunking granularity reverses with training state, structure-aware before fine-tuning and fixed-size after, so chunking optimized against a base model cannot be inherited by its fine-tuned descendant. RAFT trails SFT by 2.2 to 2.3 percentage points overall, and the deficit holds in all four reactor-type and chunking strata. The pipeline runs on one workstation with no network access at run time, and the result approaches operator-level command of engineering fundamentals without reliably achieving it.

\end{abstract}
\begin{highlights}
\item Fine-tuning lifts an open 31B model from 0 to 8 of 14 NRC licensing papers passed.
\item Eight model-retrieval configurations scored on 697 items from a closed program.
\item Preferred chunking granularity reverses once the model is fine-tuned.
\item RAFT trails plain SFT in every reactor-type and chunking stratum.
\item Whole pipeline runs air-gapped on one workstation with no run-time dependency.
\end{highlights}

\date{\today}
\maketitle
% =========================================================
% 1. INTRODUCTION
% =========================================================
\section{Introduction}\label{sec:introduction}
The safe operation of nuclear power plants depends on highly trained human operators who must apply complex nuclear engineering knowledge and procedural reasoning under safety-critical conditions and under time pressure~\cite{pakarinen2025workload,nrc_idheas_2025}. In the United States (U.S.), the Nuclear Regulatory Commission (NRC) certifies these operators through the Reactor Operator (RO) license examination~\cite{nrc_10cfr55}, a rigorous assessment spanning reactor physics, plant systems, thermodynamics, instrumentation and control, and emergency operating procedures. The qualification criteria are defined in the NRC operator licensing exam standards. To obtain a license, a candidate must receive a total score of at least 80\% on the written exam, and rounding the score up to meet this passing mark is not permitted~\cite{nrc_examstandards_2021}. The generic engineering component of that written examination is the Generic Fundamentals Examination (GFE), which is the instrument used throughout this study. Because this 80\% threshold is the same standard applied to every human RO candidate, the examination lets us assess a model against a human on identical terms.

In recent years, large language models (LLMs) have advanced rapidly in language understanding, multi-step reasoning, and information retrieval. A growing body of work uses standardized professional licensing examinations, most prominently in medicine and law. Singhal et~al.~\cite{singhal2023medpalm} showed that Med-PaLM~2 could approach expert clinician performance on questions styled after the United States Medical Licensing Examination (USMLE), and Nori et~al.~\cite{nori2023gpt4usmle} demonstrated that GPT-4 attained passing scores across all three USMLE steps without any medical-specific fine-tuning. In the legal domain, Katz et~al.~\cite{katz2024gpt4bar} reported that GPT-4 passed the Uniform Bar Examination at approximately the 90th percentile among human test-takers. Because these examinations are verified, professionally administered, and graded against fixed passing criteria, strong performance on them is meaningful evidence of domain expertise rather than superficial pattern matching, which motivates our use of the NRC RO examination as a benchmark.

Despite this progress, adoption of LLMs within the nuclear industry has remained limited. Existing applications of natural language processing in nuclear engineering have concentrated on document classification, anomaly detection, predictive maintenance, and natural-language interfaces to plant data~\cite{wang2024technical,wang2025large,qi2026synergistic,fayyaz2025natural,LEE2025103842,lin2025using}, rather than on demonstrating engineering knowledge competence on the standardized licensure assessments that define operator qualification. The obstacle common to every such deployment is the well-documented tendency of LLMs to produce fluent but factually incorrect outputs, a failure mode difficult to reconcile with reactor operations, where a single incorrect procedural recommendation can carry severe safety consequences. Beyond that shared obstacle, the medical and legal results do not transfer to this setting for three reasons specific to it. The first is scale of evidence. Medicine and law are among the most heavily represented professional domains in web-scale pretraining corpora, whereas reactor operations are documented largely in regulatory filings, vendor manuals and utility procedures that are either proprietary or sparsely indexed, so strong zero-shot performance cannot be assumed. The second is modality. A substantial fraction of licensing items are inseparable from a figure, and a text-only evaluation would silently discard exactly the items that exercise engineering reading. The third is deployment posture. Passing scores obtained from a proprietary frontier endpoint carry little operational weight in a sector where data governance and air-gapped operation are frequently prerequisites rather than preferences, so the question of interest is not whether some model can pass but whether a model that a utility could actually host can be brought to the criterion. Closing this gap requires capable base models together with fine-tuning strategies that ground generation in authoritative, domain-specific sources.

We therefore ask whether an LLM can reach a reliable standard of competence on the NRC RO license examination. To address the question we evaluate a candidate model on the exam, the same instrument applied to human candidates, and hold it to the same regulatory 80\% passing standard. Performance at or above this threshold would be concrete evidence that a model possesses operationally relevant knowledge of reactor physics and regulatory procedure, supporting its potential deployment as a digital co-pilot that augments, rather than replaces, the judgment of a licensed human operator.

Two complementary techniques are commonly used to ground general-purpose LLMs in specialized domains. The first, supervised fine-tuning (SFT)~\cite{hsieh2023distillingstep}, updates model parameters on domain-specific knowledge. Its effectiveness increases substantially when the training targets are step-by-step chain-of-thought (CoT) rationales distilled from a stronger teacher model, a strategy that transfers reasoning ability to smaller students far more efficiently than label-only supervision~\cite{magister2023teaching}. The second, retrieval-augmented generation (RAG)~\cite{lewis2020retrieval}, leaves parameters untouched and instead injects authoritative external context at inference time, augmenting the model's parametric knowledge with retrieved evidence to suppress hallucination on knowledge-intensive queries. Retrieval-augmented fine-tuning (RAFT) combines the two~\cite{zhang2024raft}, training the model on retrieved context so that it learns to attend to relevant passages while remaining robust to imperfect or distracting retrieval. A critical design parameter in all RAG-based methodologies is chunking, which dictates how source documents are segmented into discrete, retrievable units. Chunk granularity is known to affect retrieval quality~\cite{llamaindex2024chunking}, and its interaction with the model's training state has received little attention, particularly when retrieval is paired with fine-tuning.

A further complication specific to technical and regulatory domains is that the source material is intrinsically multimodal. NRC examinations and the U.S.~Department of Energy (DOE) Fundamentals Handbooks that support them contain many equations~\cite{doe_hdbk_thermo, doe_hdbk_ic, doe_hdbk_nuclear}, piping-and-instrumentation diagrams (P\&IDs), system schematics, and annotated graphs that are integral to the questions they accompany. Answering such items therefore requires a model that reasons over both text and images, not text alone. This motivates our use of an open-weight multimodal model. The instruction-tuned Gemma~4 family~\cite{gemmateam2026gemma4} provides a publicly available 31-billion-parameter variant with native image understanding, well suited to reproducible domain-adaptation studies on image-bearing regulatory content.

A final consideration constrains the choice of model. Utilities and national laboratories routinely operate under data-governance regimes that prohibit transmission of plant-related text to third-party inference endpoints, and regulators require that a deployed decision aid be auditable in the sense that its weights, its training corpus and its retrieval index can be inspected in one place. These constraints exclude closed frontier models from the deployment path irrespective of their measured accuracy. They motivate an evaluation whose subject is an open-weight model of a size that fits on a single workstation, and whose entire run-time path is free of network dependence. We adopt that posture throughout and treat it as a design requirement rather than as a convenience.

Building on these observations, this study presents a systematic benchmark of a 31-billion-parameter open-weight multimodal model (Gemma~4 31B-IT) across eight model--retrieval configurations on the NRC Reactor Operator Generic Fundamentals Examination, covering both reactor types. {The contributions are the following.}

\begin{enumerate}
\item {\textbf{A licensure-faithful benchmark:} We evaluate against the instrument the regulator actually administered and against the criterion the regulator actually applies, paper by paper and without rounding, on a census of the March administrations of a now-closed program. The unit of assessment is inherited from the standard rather than chosen for statistical convenience, and we show in Section~\ref{sec:discussion:unit} that pooled accuracy inverts the qualitative conclusion that the licensure unit supports.}
\item {\textbf{Reproducible and self-contained adaptation pipeline:} We adapt a 31B multimodal model to nuclear regulatory question answering, including image-bearing items such as P\&IDs and system schematics, through distilled CoT supervision, merged low-rank adaptation (LoRA) adapters, deterministic BM25 retrieval and an MLX inference back-end. Fine-tuning and inference both execute on one workstation, and the deployed artifact requires no proprietary model and no network access.}
\item {\textbf{A quantified account of what closes the gap:} We establish that the distance from 51.94\% out of the box to the 80\% criterion is closable without a larger model or a different architecture, that SFT with fixed-size chunking retrieval satisfies the criterion on 8 of 14 papers while no configuration without fine-tuning satisfies it on any, and that distilled CoT supervision rather than retrieval is the dominant contributor, with the two interventions markedly sub-additive.}
\item {\textbf{Two transferable design findings and one diagnostic:} We report a chunking-strategy reversal, in which structure-aware segmentation is preferred before fine-tuning and fixed-size sliding windows after, so retrieval design is contingent on training state; a consistent deficit of RAFT relative to plain SFT under matched chunking and matched inference-time retrieval, for which we identify a testable mechanism; and a difficulty diagnostic that separates intrinsic item difficulty from training-coverage gaps using only quantities a benchmark already produces.}
\end{enumerate}

% =========================================================
% 2. METHODOLOGY
% =========================================================
\section{Methodology}\label{sec:methodology}
\begin{figure*}[t]
  \centering
  % pipeline.tex -- fits \textwidth natively. Do NOT wrap in \resizebox.
% This file must contain no blank lines: a blank line becomes \par.
% Style names avoid TikZ reserved keys (out, in, at, to, node, edge, ...).
% Layout: three lanes across the top. The administered papers overlap the
% question banks, and that overlap is removed from the banks by NORMALIZED
% QUESTION TEXT matching, not by identifier. Dashed arrows carry produced
% components and supplied context; solid arrows carry the item stream.
% The teacher-side dense retriever is drawn in accent colour because the
% mismatch against the student-side sparse retriever is the mechanism
% proposed in Section 4.4.
\begin{tikzpicture}[
  font=\sffamily\small,
  stage/.style={rectangle, rounded corners=3pt, draw=black!75, line width=0.7pt,
    fill=white, text width=3.1cm, align=center, minimum height=1.5cm, inner sep=5pt},
  source/.style={stage, fill=black!8, draw=black!85},
  terminal/.style={stage, fill=black!12, draw=black!85, line width=1.0pt},
  flow/.style={-{Stealth[length=2.6mm,width=2.0mm]}, draw=black!75, line width=0.8pt},
  feed/.style={-{Stealth[length=2.6mm,width=2.0mm]}, draw=black!60, line width=0.7pt,
    dash pattern=on 2.4pt off 1.8pt},
  teach/.style={-{Stealth[length=2.6mm,width=2.0mm]}, draw=blue!55!black, line width=0.7pt,
    dash pattern=on 2.4pt off 1.8pt},
  lanelabel/.style={font=\sffamily\small\bfseries\itshape, text=black!80, inner sep=2pt},
  edgetag/.style={font=\sffamily\footnotesize, text=black!60, align=center, inner sep=2pt},
  teachtag/.style={font=\sffamily\footnotesize, text=blue!55!black, align=center, inner sep=2pt}
]
\node[lanelabel] at (2.6, 1.35) {Training corpus};
\node[lanelabel] at (8.4, 1.35) {Evaluation};
\node[lanelabel] at (13.8, 1.35) {Retrieval corpus};
\node[source] (banks) at (2.6, 0) {\textbf{NRC question banks}\\[2pt] PWR 2{,}140\\ BWR 2{,}149};
\node[source] (exams) at (8.4, 0) {\textbf{Administered papers}\\[2pt] March 2015 to 2021\\ 14 papers};
\node[source] (books) at (13.8, 0) {\textbf{DOE handbooks}\\[2pt] 7 volumes};
\node[stage] (dedup) at (2.6, -2.4) {\textbf{Deduplication}\\[2pt] by normalized\\ question text};
\node[stage] (prep) at (8.4, -2.4) {\textbf{Item preparation}\\ 700 slots, 3 marked \\ deleted by NRC};
\node[stage] (chunk) at (13.8, -2.4) {\textbf{Chunking}\\[2pt] fixed-size and\\ structure-aware};
\node[stage] (cot) at (2.6, -4.8) {\textbf{CoT distillation}\\[2pt] Gemini 3 Flash teacher};
\node[stage] (retr) at (13.8, -4.8) {\textbf{BM25 sparse index}\\[2pt] top-4, 7{,}000 chars};
\node[stage] (tune) at (2.6, -7.2) {\textbf{LoRA fine-tuning}\\[2pt] SFT and RAFT};
\node[stage] (infer) at (8.4, -7.2) {\textbf{Model inference}\\[2pt] 8 configurations\\ greedy decoding};
\node[stage] (extract) at (8.4, -9.4) {\textbf{Answer extraction}\\[2pt] cascaded matcher};
\node[terminal] (results) at (8.4, -11.4) {\textbf{Pass count and}\\ \textbf{pooled accuracy}\\[2pt] 80\% criterion};
\draw[flow] (banks) -- (dedup);
\draw[flow] (books) -- (chunk);
\draw[flow] (chunk) -- (retr);
\draw[flow] (dedup) -- node[edgetag, right] {3{,}578 items} (cot);
\draw[flow] (cot) -- node[edgetag, left] {3{,}577\\items} (tune);
\draw[flow] (exams.west) -- (4.85,0) -- (4.85,-2.4) -- (dedup.east);
\node[edgetag, anchor=west] at (4.97,-1.35) {verbatim\\ text match};

\draw[feed] (retr.west) -- (11.0,-4.8) -- (11.0,-6.0) -- node[edgetag, above, pos=0.62] {retrieved context, RAFT only} (3.4,-6.0) -- (3.4,-6.45);
\draw[feed] (tune) -- node[edgetag, above] {merged\\ adapter} (infer);
\draw[feed] (retr.south) |- node[edgetag, above, pos=0.72] {retrieved context} (infer.east);
\draw[flow, preaction={draw, line width=2.2pt, white}] (exams) -- node[edgetag, right] {700 items} (prep);
\draw[flow, preaction={draw, line width=2.2pt, white}] (prep) -- node[edgetag, right] {697 scored} (infer);
\draw[flow] (infer) -- (extract);
\draw[flow] (extract) -- (results);
\end{tikzpicture}%
 
  \caption{Pipeline of the proposed framework. The question banks and the administered papers are obtained independently from the NRC and the papers overlap the banks. Deduplication removes from the pooled banks every item whose normalized question text matches an evaluation item, leaving 3,578 training questions. One further item was dropped when the teacher API call returned an error, so 3,577 receive distilled CoT rationales and are used for low-rank fine-tuning. The DOE handbook corpus is segmented under two chunking policies and indexed for BM25 retrieval. Eight configurations are evaluated on the fourteen administered papers, each scored independently against the 80\% criterion. Three of the 700 administered slots are marked deleted in the public examination copies and are excluded, leaving 697 scored items.}
  \label{fig:pipeline}
\end{figure*}
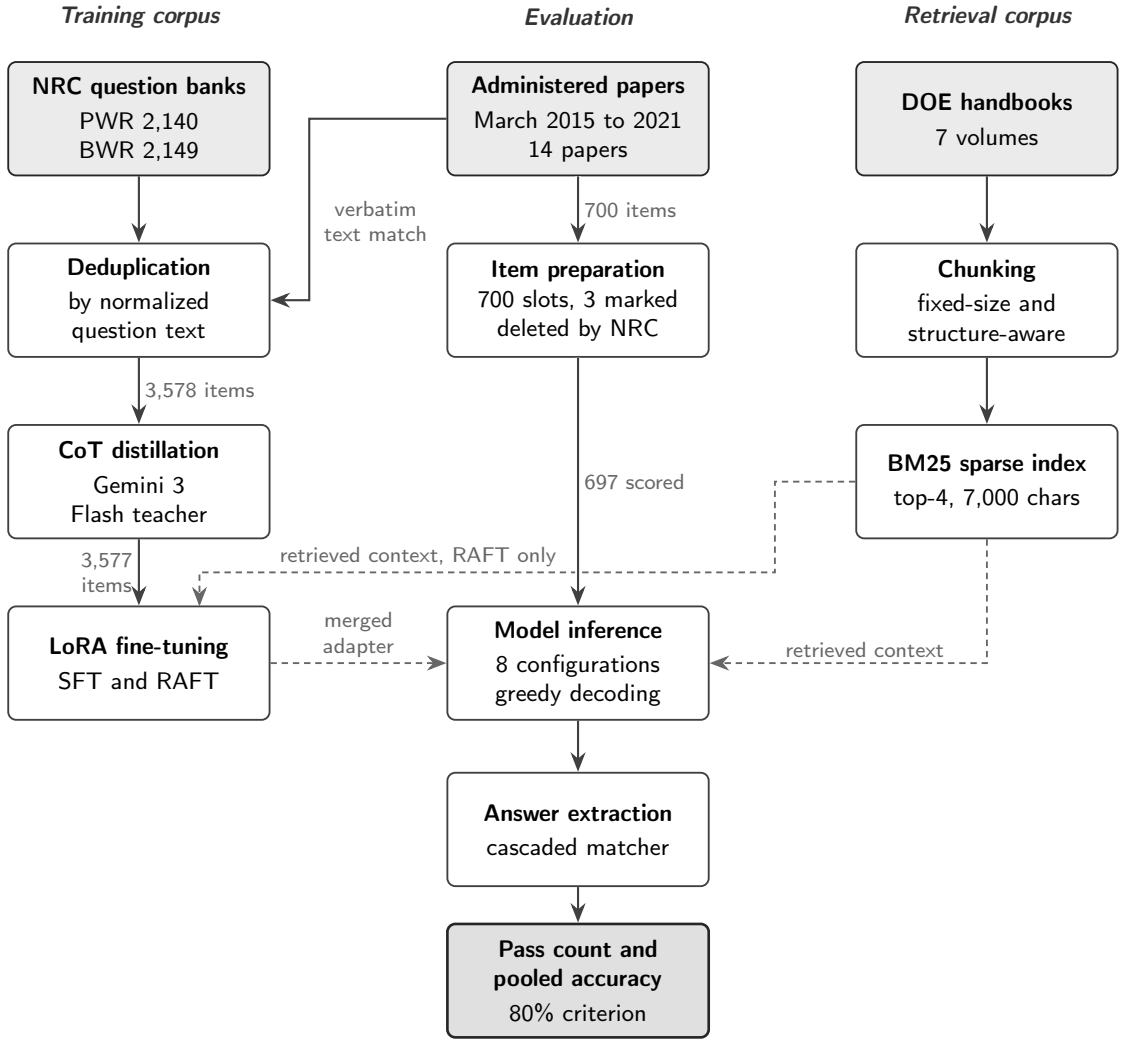

This section describes the evaluation framework, the dataset and the computational pipeline. Figure~\ref{fig:pipeline} summarizes the workflow.

% --- 2.1 ---
\subsection{Problem Formulation and Evaluation Standard}\label{sec:methodology:formulation}

The task is a closed-ended, four-choice question-answering problem drawn from the GFE. Let $X$ denote an input query, consisting of a technical scenario, the options $\{A,B,C,D\}$, and any accompanying images, and let $Y \in \{A,B,C,D\}$ denote the ground-truth answer. The objective is to maximize $P(Y \mid X)$ for unaugmented configurations and $P(Y \mid X, C)$ for retrieval-augmented configurations, where $C$ denotes external technical context retrieved from a reference corpus.

The passing criterion is fixed by regulation rather than chosen by the authors. A candidate must score at least 80\%, and rounding up to reach that mark is not permitted. This criterion applies to a single administered examination and not to an average across sittings. We therefore adopt the individual administered examination, rather than a pooled item set, as the primary unit of evaluation.

Each examination contains 50 administered item slots. Three slots are marked deleted in the publicly available examination copies and provide neither scorable stem nor answer options, \texttt{P-17\_Q14}, \texttt{B-17\_Q15}, and \texttt{B-21\_Q45}. The two 2017 papers and the 2021 BWR paper are therefore scored over 49 items. Scoring is strictly binary. The model receives 1 point for a correct answer and 0 for an incorrect one. We mirror the actual examination convention here, under which a candidate who leaves an item unanswered receives no credit for it. For a given examination $e$, the pass indicator is
\begin{equation}
\text{Pass}(e) =
\begin{cases}
1 & \text{if } \dfrac{\text{Correct answers in } e}{\text{Total items in } e} \geq 0.80 \\[4pt]
0 & \text{otherwise}
\end{cases}
\label{eq:passthreshold}
\end{equation}
so that Equation~\eqref{eq:passthreshold} evaluates to 1 only when at least 40 of 50 items are answered correctly. The requirement is 40 of 49 for the two 2017 papers and the 2021 BWR paper, and 40 of 50 for every other paper.

Our primary metric is the number of examinations passed out of the 14 administered papers, reported in Table~\ref{tbl:pass_counts}. As a secondary metric we report pooled accuracy, the percentage of correct answers across all items from all examinations combined. Pooled accuracy offers a broader statistical summary but has no direct regulatory interpretation, since candidates are never assessed on a pooled item set drawn from multiple papers.

The scope of the instrument is limited in one further respect. The GFE section does not include questions on site-specific regulations, technical specifications, or emergency operating procedures. Because full operator licensing also requires a separate site-specific written examination and an operating test, the claims in this paper are strictly limited to the generic knowledge component of licensure.

% --- 2.2 ---
\subsection{Evaluation Dataset and Training Corpus}\label{sec:methodology:dataset}

\subsubsection{Data Sources}
Three publicly available NRC and DOE sources form the experimental corpus.

The first is the pair of NRC GFE question banks for pressurized water reactors~\cite{nrc_pwr_bank_2020} and boiling water reactors~\cite{nrc_bwr_bank_2020}, the authoritative pools from which examination questions are drawn. Both are used to construct the training corpus. The PWR question bank contains 2{,}140 questions and the corresponding BWR question bank contains 2{,}149 questions. Each bank is organized across three subject areas. These are Components under Topic 191xxx, Reactor Theory under Topic 192xxx and Thermodynamics under Topic 193xxx. Each item carries a unique question identifier and a knowledge tag encoding the competency tested. A subset of items includes engineering diagrams, schematics or piping and instrumentation figures as integral parts of the question stem. These are extracted as native images and processed through the multimodal interface of the model.

The second is the set of administered examinations that forms the evaluation set, comprising every GFE administered at the March sitting from 2015 through 2021 for both reactor types~\cite{nrc_pwr_2015mar, nrc_pwr_2016mar, nrc_pwr_2017mar, nrc_pwr_2018mar, nrc_pwr_2019mar, nrc_pwr_2020mar, nrc_pwr_2021mar, nrc_bwr_2015mar, nrc_bwr_2016mar, nrc_bwr_2017mar, nrc_bwr_2018mar, nrc_bwr_2019mar, nrc_bwr_2020mar, nrc_bwr_2021mar}. Each examination consists of 50 four-choice items. This yields 350 PWR items and 350 BWR items across fourteen administrations. The March sitting is selected because it is the only sitting available in every year of the study period. The study period terminates in 2021 because the last GFE was administered in September of that year, the NRC having discontinued the program on 16 March 2022~\cite{nrc_gfe_history}. The evaluation set is therefore a census of the March administrations of a closed program rather than a sample of an ongoing series.

The third is the external retrieval corpus, built from seven volumes of the DOE Fundamentals Handbook series, which the NRC recommends as preparatory reading. They comprise \emph{Thermodynamics, Heat Transfer, and Fluid Flow} in volumes~1 to 3, \emph{Instrumentation and Control} in volumes~1 and 2 and \emph{Nuclear Physics and Reactor Theory} in volumes~1 and 2. The handbooks are agnostic to reactor type. They cover the subject areas tested for both PWR and BWR examinations. We exploit this property in Section~\ref{sec:results:asymmetry}.

\subsubsection{Train and Evaluation Split}
A deduplication procedure prevents overlap between the training corpus and the evaluation set. The two question banks are pooled into a single set of 4{,}289 items. Each pooled item and each evaluation item is reduced to a normalized signature by deleting all non-alphanumeric characters and lower-casing the remainder. Every pooled item whose signature matches that of an evaluation item is removed. Matching is therefore on question text rather than on question identifier. The procedure removed 711 items, corresponding to 472 distinct normalized evaluation signatures occurring across 573 of the 700 administered item rows, and 3{,}578 items were retained. Table~\ref{tbl:dataset_split} summarizes the composition.

The number of items removed exceeds the number of evaluation questions matched because many questions appear in both banks. Only 3{,}227 of the 4{,}289 pooled signatures are distinct, and 2{,}069 items share a signature with at least one other item. A single evaluation item therefore deletes every pooled item carrying the same text. The procedure does not remove duplication internal to the training corpus, so a question common to both reactor types contributes more than one training example.

Text matching removes an evaluation question only where its wording is reproduced up to punctuation and case. The NRC constructs each administered examination from three sources~\cite{nrc_gfe_history}. Forty items are drawn directly from the bank. Five are derived by modifying bank items through one or more significant modifications. Five are newly authored. Verbatim reuse is therefore eliminated. A derived item and its bank ancestor differ in wording and both survive. We treat the residual similarity between such pairs as a bounded but non-zero source of optimism and return to it in Section~\ref{sec:discussion:limitations}.

\begin{table}[pos=h]
\caption{Dataset composition. Deduplication removed 711 of the 4{,}289 pooled bank items, and one further item was discarded during rationale generation, leaving the training corpus shown. The training rows are stated after both removals, so differencing a bank row against its training row accounts for 712 items in total, of which 711 are deduplication removals and one is the discarded item.}
\label{tbl:dataset_split}
\begin{tabular*}{\tblwidth}{@{} LLL @{}}
\toprule
\textbf{Split} & \textbf{Source} & \textbf{Items} \\
\midrule
Question banks        & PWR & 2{,}140 \\
                      & BWR & 2{,}149 \\
Bank total            &     & 4{,}289 \\
\midrule
Training (SFT / RAFT) & PWR question bank & 1{,}797 \\
                      & BWR question bank & 1{,}780 \\
Training total        &                   & 3{,}577 \\
\midrule
Evaluation            & PWR, March 2015 to 2021 & 350 \\
                      & BWR, March 2015 to 2021 & 350 \\
Evaluation total      &                   & 700 \\
\bottomrule
\end{tabular*}
\end{table}

\subsubsection{Chain-of-Thought Distillation}\label{sec:methodology:cot}
The 3{,}578 retained items are augmented with step-by-step rationales produced by a stronger teacher model. One item produced no output because the teacher API call terminated with an error, and no retry was attempted, so that item was dropped. The corpus used for fine-tuning therefore comprises 3{,}577 examples, of which 1{,}797 originate in the PWR bank and 1{,}780 in the BWR bank. No other item was excluded at this stage. For each question the stem, the four options and the ground-truth answer are submitted to \emph{Gemini~3 Flash} together with reference passages retrieved from the handbook corpus. The teacher is instructed to explain step by step why the indicated option is correct using engineering principles present in the supplied references. It concludes with an explicit \texttt{"Therefore, the correct answer is (X)"} statement. The rationale is stored as a two-turn dialog. The user turn is the question with options and the assistant turn is the distilled rationale. For items containing diagrams the images are passed to the teacher alongside the text so that the rationale can reference visual content. The same images are retained in the training sample and routed through the vision encoder of the student.

The teacher-side retriever differs from the retriever used at evaluation. The passages supplied to the teacher are retrieved by a dense index built once over the same handbook texts. It uses FAISS over \texttt{all-MiniLM-L6-v2} embeddings and returns the top three passages. The evaluation-time retriever is sparse and returns the top four chunks, as described in Section~\ref{sec:methodology:rag}. For the SFT configurations this difference is inconsequential. SFT training examples contain no retrieved context. The retriever of the teacher therefore affects only the content of the rationale that the student learns to reproduce from parametric knowledge. For RAFT the situation differs. The training example pairs a sparse-retrieved context block with a rationale written against a differently retrieved passage set. The supervision signal may therefore assert facts that are absent from the context on which the student is simultaneously conditioned. We record this asymmetry here because it bears on the interpretation of the RAFT results in Section~\ref{sec:discussion:raft}. It was not a designed manipulation but a property of the pipeline that we identify retrospectively.

\subsubsection{Evaluation Protocol}\label{sec:methodology:eval_protocol}
Each test item is presented as a structured prompt containing the question stem, the four labeled options and the corresponding image where a figure is present. Images are passed through the multimodal interface as native pixel inputs. No answer choices are masked or modified.

For retrieval-augmented configurations the full item text, comprising the question stem and all four options, is submitted as a query against the chunked handbook corpus before generation. The top four chunks are jointly capped at 7{,}000 characters and prepended as a \texttt{[Retrieved Reference Context]} block. The prompt template is shared across all retrieval-augmented conditions.

All generation is deterministic at temperature zero with greedy decoding and up to 4{,}096 new tokens. The free-form response is post-processed by a cascaded answer extractor applying three strategies in order of priority. The first matches explicit answer phrases such as \textit{``the correct answer is X''} or \textit{``Final answer: X''}. The second takes the last parenthesized option token within the final 300 characters. The third takes a terminal standalone option token. Responses yielding no extractable option are scored as incorrect.

Three item slots are excluded from all conditions because the public examination copies mark them as deleted and provide neither a scorable stem nor answer options. These are \texttt{P-17\_Q14}, \texttt{B-17\_Q15}, and \texttt{B-21\_Q45}. The two 2017 papers and 2021 BWR paper are therefore scored over 49 items, and the remaining papers over 50. Section~\ref{sec:results:scoring} reports a sensitivity analysis that restores all three deleted slots and treats them as incorrect.

For the statistical analysis, pooled accuracy is computed over 697 scored items with 349 PWR and 348 BWR items. For each configuration we report the point estimate with a Wilson score 95\% confidence interval for a binomial proportion. Individual examinations comprise 49 or 50 scored items. The corresponding interval is approximately 11 percentage points in each direction. We therefore do not interpret the pass or failure of any single examination as evidence about that examination. Per-examination results are confined to the aggregated count $\sum_e \text{Pass}(e)$. That count is reported over the complete set of fourteen administrations without selection. No examination or subgroup was chosen after inspection of results.

Two properties of this design govern how the intervals should be read. First, the Wilson interval is a marginal interval on a single proportion. It is the appropriate instrument for the position of one configuration relative to the fixed 80\% criterion, which is the comparison the regulator defines, and we use it only for that purpose. Second, every configuration is evaluated on the identical item set under identical decoding settings, so any comparison between two configurations is paired at the item level. The variance of a paired difference depends on the discordant items alone and is in general far smaller than the sum of the marginal variances, with the consequence that overlapping marginal intervals do not license the inference that two configurations are indistinguishable. We are explicit about this because the intervals for the four fine-tuned configurations do overlap substantially, and a reader who treats those intervals as a test of difference would draw the wrong conclusion in both directions.

We nonetheless refrain from reporting nominal $p$ values for the between-configuration contrasts, because the eight configurations generate twenty-eight pairwise comparisons and no multiplicity correction was pre-registered. In place of a nominal test we adopt a stratum sign criterion, which is conservative and pre-specified. A difference is described as robust in direction only when it holds in every stratum in which it can be evaluated, where the strata are formed by crossing reactor type with chunking strategy. Four such strata are available for the SFT against RAFT contrast. Under the null hypothesis that the sign of the difference is a fair coin in each stratum, unanimity across four strata corresponds to a one-sided probability of $2^{-4} = 0.0625$. This criterion supports statements about the direction of an effect and supports no statement about its magnitude, and we confine our claims accordingly throughout Sections~\ref{sec:results} and~\ref{sec:discussion}. Where a paired item-level test is the natural next instrument, we identify it as such in Section~\ref{sec:discussion:limitations} rather than reporting it post hoc.

% --- 2.3 ---
\subsection{Corpus Segmentation}\label{sec:methodology:chunking}
The external knowledge base $\mathcal{K}$ is constructed from the seven handbook volumes. Two segmentation strategies are compared. Both operate directly on the source documents.

\subsubsection{Fixed-Size Sliding-Window Chunking}
A sliding-window function partitions each handbook into overlapping chunks $c_j \in \mathcal{K}$. Each chunk is a fixed-length span of characters, and each chunk shares its trailing 200 characters with the next chunk's leading 200 characters, as illustrated below.
\begin{equation}
|c_j| = N_{\text{size}}, \qquad |c_j \cap c_{j+1}| = N_{\text{overlap}}
\end{equation}
with $N_{\text{size}}=1000$ characters and $N_{\text{overlap}}=200$ characters, so consecutive chunks overlap by 20\% of their length. The overlap causes any given assertion in the source text to appear in more than one retrievable unit. This is the property Section~\ref{sec:discussion:chunking} appeals to in explaining why the scheme suits the fine-tuned model. Section~\ref{sec:results:chunking} examines the consequences of the redundancy. Pages dominated by exercise or problem content are discarded before chunking so that the corpus contains only expository theory text. A page is discarded if it contains four or more leading tokens of the form \texttt{Eq.}, \texttt{Ex.}, \texttt{Q.}, \texttt{Prob.} or numbered problem listings. A page is also discarded if the keywords \emph{problems}, \emph{exercises} or \emph{homework} appear within its first 500 characters.

\subsubsection{Structure-Aware Chunking}\label{sec:methodology:structure}
The second strategy partitions documents along their inherent structural boundaries rather than at arbitrary character offsets. The procedure proceeds in three stages.
\begin{enumerate}
\item \textbf{Base-style estimation:} The font-size histogram of the first fifteen pages of each handbook is computed. The modal font size is taken as the body-text size $f_{\text{base}}$.
\item \textbf{Heading detection:} A text span is a candidate heading if two conditions hold. Its rendered font size must satisfy $f_{\text{span}} \geq f_{\text{base}} + 0.8\,\text{pt}$. It must also either match a programmatically extracted table-of-contents entry or begin with a numbered prefix of the form $\{n_1\}.\{n_2\}\dots$ followed by a capitalized word. White-on-white spans are discarded to avoid spurious matches against running headers.
\item \textbf{Section accumulation:} Body text between consecutive headings is accumulated into a single chunk. Figures whose captions begin with \texttt{Figure}, \texttt{Fig.}, \texttt{FIGURE} or \texttt{FIG.} are extracted as 200~DPI rasters from the region above the caption and attached to the enclosing section chunk.
\end{enumerate}
The same problem-page filter is applied before accumulation. The resulting chunks are of variable length with a mean of approximately 1{,}435 characters. They respect the original organization of the source handbooks. We use the term \emph{structure-aware chunking} deliberately to distinguish this scheme from embedding-based semantic chunking, which we do not employ. Segmentation is driven entirely by typographic and table-of-contents cues in the source documents. No embedding model is involved. The procedure is fully deterministic given the sources.

% --- 2.4 ---
\subsection{Fine-Tuning Paradigms}\label{sec:methodology:finetuning}

\subsubsection{Supervised Fine-Tuning}\label{sec:methodology:sft}
Parameter-efficient fine-tuning uses LoRA~\cite{hu2022lora} on the 3{,}577 distilled training items of Section~\ref{sec:methodology:cot}. Only LoRA parameters are updated. Base weights remain frozen. LoRA modules are restricted to the language-model side of the architecture. The vision tower and the multi-modal projector remain frozen. This preserves the pretrained image encoder while permitting language-side specialization. Target modules are all \texttt{nn.Linear} leaves matching $\{$\texttt{q\_proj}, \texttt{k\_proj}, \texttt{v\_proj}, \texttt{o\_proj}, \texttt{gate\_proj}, \texttt{up\_proj}, \texttt{down\_proj}$\}$ outside the vision and projector subgraphs.

Training minimizes the standard next-token prediction loss, but computed only over the assistant's rationale, not over the question itself. Concretely, the model is shown the question $X$ and the teacher's distilled rationale, one token at a time, and at each position $t$ it is penalized in proportion to how much probability it assigned to the wrong next token $y_t$ given everything before it. The user's question tokens are excluded from this penalty by masking them at label $-100$, a standard convention that tells the loss function to skip them, so the model is trained to produce the rationale but not to reproduce the question.

\begin{equation}
\mathcal{L}_{\text{SFT}}(\theta) = -\sum_{t=1}^{T} \log P_{\theta}\bigl(y_t \mid y_{<t},\, X\bigr)
\label{eq:sftloss}
\end{equation}

Here $\theta$ denotes the trainable LoRA parameters, $T$ is the number of rationale tokens, $y_t$ is the $t$-th rationale token, and $y_{<t}$ denotes every rationale token generated before it. Minimizing $\mathcal{L}_{\text{SFT}}$ is equivalent to maximizing the probability the model assigns to the teacher's rationale, token by token, conditioned on the question. Hyperparameters appear in Table~\ref{tbl:training_hparams}. Two safeguards protect against silent training failure. The trainable parameter count is asserted non-zero immediately after LoRA injection. A callback aborts training if the gradient norm is exactly zero across the first five logged steps. The training corpus pools PWR and BWR items. A single fine-tuned model is therefore produced and evaluated on both reactor types.

\begin{table}[pos=h]
\caption{LoRA hyperparameters. Identical settings are used for SFT and the two RAFT variants. Only the training corpus differs.}\label{tbl:training_hparams}
\begin{tabular*}{\tblwidth}{@{} LL @{}}
\toprule
\textbf{Hyperparameter} & \textbf{Value} \\
\midrule
Base model                    & \texttt{google/gemma-4-31b-it} \\
LoRA rank ($r$)               & 16 \\
LoRA scaling ($\alpha$)       & 32 \\
LoRA dropout                  & 0.05 \\
Target modules                & q/k/v/o\_proj, gate/up/down\_proj \\
                              & (language model only) \\
Frozen subgraphs              & vision tower, multi-modal projector \\
Epochs                        & 3 \\
Per-device batch size         & 1 \\
Gradient accumulation steps   & 8 \\
Effective batch size          & 8 \\
Learning rate                 & $2 \times 10^{-4}$ \\
LR scheduler                  & cosine \\
Warmup ratio                  & 0.05 \\
Weight decay                  & 0.01 \\
Maximum sequence length       & 2{,}048 \\
Precision                     & bfloat16 \\
Optimizer                     & AdamW \\
Gradient checkpointing        & enabled \\
\bottomrule
\end{tabular*}
\end{table}

\subsubsection{Retrieval-Augmented Fine-Tuning}\label{sec:methodology:raft}
RAFT trains the model to operate in the presence of imperfect retrieval. Each training example is augmented with context produced by the retrieval pipeline used at inference time. Training-time and test-time context distributions therefore match. For every training question the top four chunks are jointly capped at 7{,}000 characters and prepended under the block format used at evaluation. The assistant target remains the distilled rationale of Section~\ref{sec:methodology:cot}. The RAFT objective is identical to the SFT objective of Equation~\eqref{eq:sftloss}, with one addition. The retrieved context block $C^{\text{retr}}$ is also visible to the model when it predicts each rationale token. The model is therefore trained to use that context rather than to ignore it.
\begin{equation}
\mathcal{L}_{\text{RAFT}}(\theta) = -\sum_{t=1}^{T} \log P_{\theta}\bigl(y_t \mid y_{<t},\, X,\, C^{\text{retr}}\bigr)
\end{equation}
where $C^{\text{retr}}$ is the block of retrieved reference passages described below, concatenated with the question exactly as it appears at inference time. Two variants are trained, one per chunking strategy. Each is evaluated with the matching strategy so that chunking statistics are consistent between training and evaluation. All other hyperparameters are identical to SFT. RAFT presumes a retrieval component. The \emph{No Retrieval} condition is therefore not applicable to it.

Section~\ref{sec:methodology:cot} noted that the rationale targets were generated against dense-retrieved passages while $C^{\text{retr}}$ is sparse-retrieved. The context distribution is matched between training and inference but the supervision is not matched to the context. We treat this as an explanatory hypothesis rather than a controlled variable.

% --- 2.5 ---
\subsection{Retrieval at Inference}\label{sec:methodology:rag}

All retrieval-augmented configurations use sparse lexical retrieval with BM25~\cite{robertson2009probabilistic}. BM25 scores a chunk against a query by rewarding query terms that occur in the chunk, weighted by how rare each term is across the corpus, so that terms occurring almost everywhere contribute little. For a query $Q$ tokenized into distinct terms and a candidate chunk $D$,
\begin{equation}
\text{BM25}(Q, D) = \sum_{q \in Q} \text{IDF}(q)\,
\frac{(k_1 + 1)\, f(q, D)}{f(q, D) + K(D)}
\end{equation}

\begin{equation}
K(D) = k_1\!\left(1 - b + b\,|D|/\overline{d}\right)
\label{eq:bm25k}
\end{equation}
Here $f(q, D)$ is the frequency of term $q$ in $D$, $|D|$ is the length of $D$ in tokens, and $\overline{d}$ is the mean chunk length across the corpus, so that Equation~\eqref{eq:bm25k} normalizes for chunk length. Terms repeated within a query contribute only once. The inverse document frequency is $\text{IDF}(q) = \ln\!\left(1 + (N - n_q + 0.5)/(n_q + 0.5)\right)$, where $N$ is the number of indexed chunks and $n_q$ the number of chunks containing $q$. It is larger for terms confined to few chunks. Such terms carry more discriminating power. The fraction saturates in $f(q, D)$. Each additional occurrence therefore raises the score by a diminishing amount, so that a chunk mentioning a term ten times is not scored ten times as relevant as one mentioning it once. The term $K(D)$ penalizes long chunks that accumulate matches by containing more text. The constants $k_1 = 1.5$ and $b = 0.75$ are the standard defaults governing, respectively, the rate of saturation and the strength of the length penalty.

The retrieval query is the full item text, comprising the question stem together with all four answer options. Queries and chunks are lower-cased and tokenized with an alphanumeric regular expression. No stemming, stopword removal or query expansion is applied. Chunks shorter than 80 characters are excluded from the index, as they are typically headers or page artifacts rather than substantive content. The four highest-scoring chunks are concatenated in rank order under a cumulative budget of 7{,}000 characters. Internal whitespace is normalized. The chunk that exceeds the budget is truncated at the character level. BM25 is used in place of a dense retriever because lexical retrieval is fully reproducible from the corpus alone, without any dependence on a particular embedding model or its version.

% --- 2.6 ---
\subsection{Computational Framework}\label{sec:methodology:framework}
All experiments run on Apple Silicon hardware with an M5 Max chip and 128~GB of unified memory. The entire pipeline is therefore executable within an air-gapped environment on commodity workstation hardware.

Fine-tuning executes under PyTorch with Hugging Face \texttt{transformers} and \texttt{peft} on the Metal Performance Shaders device. Training uses bfloat16 with gradient checkpointing. For inference, each LoRA module is merged into the base parameters on CPU. This avoids additional memory pressure on the Metal device. The resulting standalone \texttt{safetensors} model is converted to the Apple MLX format~\cite{mlx2023}. Evaluation uses \texttt{mlx-vlm}, which exhibits substantially more stable memory reclamation.

All weights and activations remain in bfloat16 throughout fine-tuning, merging, conversion and inference. Decoding settings follow Section~\ref{sec:methodology:eval_protocol} and are identical across all eight conditions. Accuracy differences are therefore attributable to the model and the retrieval pipeline rather than to decoding stochasticity.

% --- 2.7 ---
\subsection{Experimental Configurations}\label{sec:methodology:experiments}
Table~\ref{tbl:experimental_matrix} summarizes the eight configurations. They cross three model states of Base, SFT and RAFT with three retrieval conditions of No Retrieval, RAG-Fixed and RAG-Structure-Aware. Every configuration is evaluated on all fourteen administered examinations.

\begin{table*}[width=\textwidth,cols=4,pos=h]
\caption{Experimental matrix. Checkmarks indicate configurations evaluated in this study. RAFT requires retrieval by design. Each RAFT variant is paired at evaluation time with the chunking strategy used during its training.}\label{tbl:experimental_matrix}
\begin{tabular*}{\tblwidth}{@{\extracolsep{\fill}} LLLL @{}}
\toprule
\textbf{Model State} & \textbf{No Retrieval} & \textbf{RAG (Fixed)} & \textbf{RAG (Structure-Aware)} \\
\midrule
\textbf{Base} & \checkmark & \checkmark & \checkmark \\
\textbf{SFT}  & \checkmark & \checkmark & \checkmark \\
\textbf{RAFT} & N/A        & \checkmark & \checkmark \\
\bottomrule
\end{tabular*}
\end{table*}

% =========================================================
% 3. RESULTS
% =========================================================
\section{Results}\label{sec:results}
This section reports the performance of all eight configurations on the fourteen administered papers. The primary quantity is the number of examinations satisfying the criterion of Equation~\eqref{eq:passthreshold}. Pooled accuracy is reported as a secondary summary. All accuracies follow the strict convention of Section~\ref{sec:methodology:formulation}, under which an unextractable response is scored as incorrect and the denominator is fixed. Section~\ref{sec:results:scoring} reports the alternative convention. Every evaluation item was removed from the training corpus by the deduplication of Section~\ref{sec:methodology:dataset}. Reported scores therefore measure generalization to unseen questions rather than recall of the question bank.

\subsection{Examinations Passed}\label{sec:results:passed}
Table~\ref{tbl:pass_counts} reports the number of administered examinations on which each configuration met the criterion. The separation between the fine-tuned and unadapted groups is categorical rather than graded: the two groups do not overlap on this metric at any point, and no ordering of the three base configurations by pooled accuracy would place any of them above the weakest fine-tuned configuration. Two outcomes hold across all configurations. No configuration without fine-tuning met the criterion on any of the fourteen examinations, whether or not retrieval was supplied. The best such configuration peaked at 76.0\% on its strongest paper. The strongest fine-tuned configuration was SFT with fixed-size chunking retrieval. It met the criterion on 8 of 14 examinations and on more examinations than any alternative. The ordering among fine-tuned configurations follows the pooled ordering of Section~\ref{sec:results:overall} but separates them more sharply.

\begin{table}[pos=t]
\caption{Number of administered examinations meeting the 80\% criterion out of seven per reactor type and fourteen in total. Per-examination scores are given in Table~\ref{tab:full_matrix}, in which the dagger count of each row reproduces the corresponding row of this table.}\label{tbl:pass_counts}
\begin{tabular*}{\tblwidth}{@{\extracolsep{\fill}} LLLL @{}}
\toprule
\textbf{Configuration} & \textbf{PWR} & \textbf{BWR} & \textbf{Total} \\
\midrule
Base, no retrieval            & 0/7 & 0/7 & 0/14 \\
Base + RAG (fixed)            & 0/7 & 0/7 & 0/14 \\
Base + RAG (structure-aware)  & 0/7 & 0/7 & 0/14 \\
SFT, no retrieval             & 3/7 & 1/7 & 4/14 \\
\textbf{SFT + RAG (fixed)}    & \textbf{3/7} & \textbf{5/7} & \textbf{8/14} \\
SFT + RAG (structure-aware)   & 3/7 & 2/7 & 5/14 \\
RAFT (fixed)                  & 2/7 & 3/7 & 5/14 \\
RAFT (structure-aware)        & 3/7 & 1/7 & 4/14 \\
\bottomrule
\end{tabular*}
\end{table}

\subsection{Pooled Accuracy}\label{sec:results:overall}
Table~\ref{tbl:main_results} reports pooled accuracy by reactor type under the strict convention. Two reference points bound the scale. Uniform random selection over four options gives 25\%, and the regulatory criterion is 80\%, so the base model at 51.94\% has traversed slightly under half the distance between chance and the criterion without any domain adaptation. The interventions studied here are therefore closing the harder half. The accuracy reaches the criterion on the PWR items alone. SFT with fixed-size chunking retrieval scores 80.23\% on PWR, which corresponds to 280 of 349 items and clears the mark by a single item, since 279 correct answers would give 79.94\%. The same configuration reaches 79.31\% on BWR and 79.77\% pooled, both below the criterion. The Wilson interval on the pooled figure spans the threshold. The pooled comparison is therefore not decisive in either direction. The per-examination count of Section~\ref{sec:results:passed} is the quantity on which we rest the substantive claim.

\begin{table*}[width=\textwidth,cols=5,pos=h]
\caption{Pooled accuracy in percent across all administered examinations by reactor type, under the strict convention. Denominators are 349 PWR items, 348 BWR items and 697 pooled. The interval is the Wilson score 95\% interval on the combined figure.}\label{tbl:main_results}
\begin{tabular*}{\tblwidth}{@{\extracolsep{\fill}} Lcccc @{}}
\toprule
\textbf{Configuration} & \textbf{PWR} & \textbf{BWR} & \textbf{Combined} & \textbf{95\% CI} \\
\midrule
Base, no retrieval           & 52.15 & 51.72 & 51.94 & [48.2, 55.6] \\
Base + RAG (fixed)           & 59.31 & 65.23 & 62.27 & [58.6, 65.8] \\
Base + RAG (structure-aware) & 61.03 & 68.10 & 64.56 & [60.9, 68.0] \\
SFT, no retrieval            & 76.50 & 74.71 & 75.61 & [72.3, 78.7] \\
\textbf{SFT + RAG (fixed)}   & \textbf{80.23} & 79.31 & 79.77 & [76.6, 82.6] \\
SFT + RAG (structure-aware)  & 78.22 & 77.01 & 77.62 & [74.4, 80.6] \\
RAFT (fixed)                 & 77.08 & 77.87 & 77.47 & [74.2, 80.4] \\
RAFT (structure-aware)       & 77.08 & 73.85 & 75.47 & [72.1, 78.5] \\
\bottomrule
\end{tabular*}
\end{table*}

Distilled CoT fine-tuning is the largest single contributor to accuracy. Holding the retrieval condition fixed, SFT raises combined accuracy from 51.94\% to 75.61\% without retrieval for a gain of 23.7 percentage points. Under fixed-size chunking it raises accuracy from 62.27\% to 79.77\% for a gain of 17.5 points. Under structure-aware chunking it raises accuracy from 64.56\% to 77.62\% for a gain of 13.1 points. The fine-tuned model without retrieval reaches 75.61\%. This exceeds the best base configuration at 64.56\% by 11.1 points.

Retrieval provides a consistent secondary benefit that diminishes as fine-tuning increases. On the base model retrieval adds 10.3 points under fixed-size chunking and 12.6 points under structure-aware chunking. On the fine-tuned model the increment falls to 4.2 points and 2.0 points respectively. The two interventions are individually large but jointly sub-additive, and the comparison is made within a single chunking condition so that the addends and the target occupy the same design cell. Under fixed-size chunking, fine-tuning alone raises accuracy by 23.6 points and retrieval alone raises it by 10.3 points, for an additive prediction of 34.0 points. The observed gain from the base model to SFT with fixed-size retrieval is 27.8 points, a shortfall of 6.2 points against that prediction. The same calculation under structure-aware chunking gives an additive prediction of 36.3 points against an observed 25.7, a shortfall of 10.6 points. Fine-tuning on bank questions therefore supplies implicitly a substantial portion of the handbook knowledge that retrieval would otherwise contribute, and the overlap is larger for the segmentation that carries more expository context per chunk.

\subsection{Retrieval Gain by Reactor Type}\label{sec:results:asymmetry}
Retrieval benefits the BWR items more than the PWR items, and the direction holds in all four available comparisons. On the base model, fixed-size chunking adds 7.2 points on PWR against 13.5 points on BWR, and structure-aware chunking adds 8.9 points against 16.3 points. On the fine-tuned model the two gaps narrow to 0.9 and 0.6 points.

The retrieval corpus offers no mechanism for the pattern. The seven handbook volumes are agnostic to reactor type and cover the subject areas tested on both examinations, so neither reactor type is better served by the retrieved text. Nor is the pattern an artefact of the base rates, which differ by 0.4 points between the two reactor types before retrieval is supplied. We report the asymmetry without an explanation. Any single one of the four differences is within sampling error with 349 PWR and 348 BWR items, so we do not claim a reactor-type effect.  Locating the pattern would require a decomposition of accuracy by topic rather than by reactor type, which we leave to future work.

{\subsection{Text-Only and Image-Bearing Items}\label{sec:results:multimodal}
The instrument is multimodal by construction, and Section~\ref{sec:introduction} rests part of the motivation for an image-capable model on that fact. This section decomposes accuracy along the modality of the item so that the claim is tested rather than assumed.}
{Of the 697 scored items, 136 carry a figure integral to the stem and 561 are text-only, so image-bearing items are 19.5\% of the evaluation set. The fraction is close to balanced between reactor types, at 67 of 349 PWR items and 69 of 348 BWR items. It is not balanced across subject areas. Components items carry a figure in 29.7\% of cases, against 6.1\% for Reactor Theory and 16.9\% for Thermodynamics, which reflects the reliance of component questions on piping and instrumentation diagrams and system schematics. The training corpus is comparably composed, with 652 of the 3{,}577 distilled examples carrying at least one image. The three items excluded in Section~\ref{sec:methodology:eval_protocol} are not assigned a modality because the public examination copies mark them as deleted and provide no scorable stems or answer options.}

{Table~\ref{tbl:modality} reports accuracy on the two strata for all eight configurations. The gap is defined as text-only accuracy minus image-bearing accuracy, so a positive value indicates that figure-bearing items are the harder stratum.}

\begin{table}[pos=h]
\caption{{Accuracy in percent on text-only and image-bearing items under the strict convention. Denominators are 561 text-only and 136 image-bearing items. The gap is text-only accuracy minus image-bearing accuracy in percentage points, so a positive gap indicates that figure-bearing items are harder.}}\label{tbl:modality}
\begin{tabular*}{\tblwidth}{@{\extracolsep{\fill}} Lccc @{}}
\toprule
\textbf{Configuration} & \textbf{Text-only} & \textbf{Image} & \textbf{Gap} \\
\midrule
Base, no retrieval            & 47.24 & 71.32 & $-$24.1 \\
Base + RAG (fixed)            & 62.21 & 62.50 & $-$0.3 \\
Base + RAG (structure-aware)  & 62.92 & 71.32 & $-$8.4 \\
SFT, no retrieval             & 78.07 & 65.44 & $+$12.6 \\
\textbf{SFT + RAG (fixed)}    & \textbf{82.53} & \textbf{68.38} & \textbf{$+$14.1} \\
SFT + RAG (structure-aware)   & 79.68 & 69.12 & $+$10.6 \\
RAFT (fixed)                  & 79.50 & 69.12 & $+$10.4 \\
RAFT (structure-aware)        & 78.79 & 61.76 & $+$17.0 \\
\bottomrule
\end{tabular*}
\end{table}

{The base model already separates the two strata, scoring 47.24\% on text-only items against 71.32\% on image-bearing items, a gap of $-$24.1 points. The negative gap indicates that image-bearing items are not the harder stratum for the base model. It cannot on its own be attributed to the vision pathway, because modality is not randomly assigned and the two strata contain different questions.}

{That change is the principal finding of this section. Section~\ref{sec:methodology:sft} restricts LoRA injection to the language model and freezes the vision tower and the multi-modal projector.  Fine-tuning reverses rather than narrows the observed gap. Under the no-retrieval condition, the gap moves from $-24.1$ points for the base model to $+12.6$ points after SFT, a change of 36.7 points. Under SFT with fixed-size retrieval, text-only accuracy is 82.53\% and image-bearing accuracy is 68.38\%, giving a gap of $+14.1$ points. Positive gaps are also observed for SFT with structure-aware retrieval and for the two RAFT configurations, at $+10.6$, $+10.4$ and $+17.0$ points, respectively. Because the vision tower and multi-modal projector are frozen, this reversal cannot reflect an improvement in the visual encoder itself. It shows that language-side adaptation changes the relative performance of the two item strata. However, it does not isolate the causal contribution of the images because the two strata contain different questions. All 136 image-bearing items were evaluated with their figures supplied, so their prompts retained the visual information required by the original examination.}

{Retrieval interacts with modality in one further respect that follows from the pipeline rather than from the model. The retrieved block is assembled from chunk text under a 7{,}000-character budget and is supplied to the model as text, so a figure attached to a structure-aware chunk during segmentation does not reach the vision encoder at inference. Retrieval therefore supplies no visual evidence to an image-bearing item, and the increment it delivers on that stratum is bounded by whatever the surrounding expository text contributes. On the base model, fixed-size retrieval adds 15.0 points on text-only items but reduces image-bearing accuracy by 8.8 points, while structure-aware retrieval adds 15.7 points on text-only items and 0.0 points on image-bearing items. Supplying the chunk-attached rasters through the multimodal interface is a modification the present pipeline does not implement and which we identify in Section~\ref{sec:conclusion} as a direct extension.}

{Two cautions bound this decomposition. Modality is not randomly assigned across the item set, and the concentration of figures in Components items means that any modality contrast is partly a subject-area contrast; separating the two would require the topic-level decomposition identified in Section~\ref{sec:results:asymmetry}. At 136 image-bearing items the Wilson interval on that stratum is approximately $\pm$8.0 points, so we read the direction of the change in the gap and not its magnitude, in accordance with the criterion set out in Section~\ref{sec:methodology:eval_protocol}.}

\subsection{Chunking-Strategy Reversal}\label{sec:results:chunking}

The preferred chunking strategy reverses with the training state of the model. The reversal holds without exception across the comparisons in Table~\ref{tbl:chunking}.

\begin{table}[pos=h]
\caption{Combined accuracy in percent by chunking strategy and model state, under the strict convention. The preferred strategy inverts after fine-tuning.}\label{tbl:chunking}
\begin{tabular*}{\tblwidth}{@{} LLLL @{}}
\toprule
\textbf{Model state} & \textbf{Fixed} & \textbf{Structure-aware} & \textbf{Preferred} \\
\midrule
Base + RAG & 62.27 & 64.56 & Structure-aware \\
SFT + RAG  & 79.77 & 77.62 & Fixed \\
RAFT       & 77.47 & 75.47 & Fixed \\
\bottomrule
\end{tabular*}
\end{table}

Structure-aware chunking is preferred for the base model at 64.56\% against 62.27\%. Fixed-size chunking is preferred for both fine-tuned paradigms. SFT reaches 79.77\% against 77.62\% and RAFT reaches 77.47\% against 75.47\%. The reversal is reproduced separately within the BWR items. On the PWR items it holds for the SFT family, at 80.23\% against 78.22\%, but the two RAFT variants are exactly tied at 77.08\%, corresponding to 269 of 349 items in both cases. Section~\ref{sec:discussion:chunking} interprets the pattern.

\subsection{RAFT Against SFT}\label{sec:results:raft}
RAFT underperforms the corresponding SFT configuration under matched chunking and matched inference-time retrieval. It reaches 77.47\% against 79.77\% under fixed-size chunking, a deficit of 2.3 points, and 75.47\% against 77.62\% under structure-aware chunking, a deficit of 2.2 points. The deficit is close to constant across chunking strategies and appears in both reactor types, at 3.2 and 1.1 points on PWR and 1.4 and 3.2 points on BWR. Because the two paradigms share identical rationale targets and identical retrieval at inference time, and differ only in whether retrieved context is present during training, the deficit is attributable to retrieval conditioning during fine-tuning.

\begin{table*}[htbp]
\centering
\caption{Accuracy in percent of all eight configurations on each administered examination, under the strict convention. The two 2017 papers and the 2021 BWR paper are scored over 49 items and all others over 50. Daggers mark examinations meeting the 80\% criterion.}
\label{tab:full_matrix}
\begin{tabular*}{\tblwidth}{@{}L@{\extracolsep{\fill}}CCCCCCCC@{}}
\toprule
Configuration & 2015 & 2016 & 2017 & 2018 & 2019 & 2020 & 2021 & Total \\
\midrule
\multicolumn{9}{@{}l}{\textit{(a) PWR}} \\
Base, no retrieval             & 56.0 & 50.0 & 53.1 & 46.0 & 56.0 & 48.0 & 56.0 & 52.15 \\
Base + RAG (fixed)             & 68.0 & 52.0 & 53.1 & 60.0 & 64.0 & 62.0 & 56.0 & 59.31 \\
Base + RAG (structure-aware)   & 64.0 & 46.0 & 67.3 & 64.0 & 64.0 & 62.0 & 60.0 & 61.03 \\
SFT, no retrieval              & 76.0 & 82.0$^\dagger$ & 73.5 & 66.0 & 84.0$^\dagger$ & 72.0 & 82.0$^\dagger$ & 76.50 \\
SFT + RAG (fixed)              & 88.0$^\dagger$ & 74.0 & 79.6 & 78.0 & 80.0$^\dagger$ & 74.0 & 88.0$^\dagger$ & 80.23 \\
SFT + RAG (structure-aware)    & 84.0$^\dagger$ & 78.0 & 71.4 & 72.0 & 82.0$^\dagger$ & 70.0 & 90.0$^\dagger$ & 78.22 \\
RAFT (fixed)                   & 78.0 & 74.0 & 77.6 & 70.0 & 82.0$^\dagger$ & 74.0 & 84.0$^\dagger$ & 77.08 \\
RAFT (structure-aware)         & 74.0 & 70.0 & 79.6 & 72.0 & 82.0$^\dagger$ & 80.0$^\dagger$ & 82.0$^\dagger$ & 77.08 \\
\midrule
\multicolumn{9}{@{}l}{\textit{(b) BWR}} \\
Base, no retrieval             & 48.0 & 48.0 & 46.9 & 70.0 & 56.0 & 42.0 & 51.0 & 51.72 \\
Base + RAG (fixed)             & 60.0 & 64.0 & 73.5 & 66.0 & 72.0 & 60.0 & 61.2 & 65.23 \\
Base + RAG (structure-aware)   & 60.0 & 70.0 & 63.3 & 76.0 & 74.0 & 58.0 & 75.5 & 68.10 \\
SFT, no retrieval              & 64.0 & 76.0 & 79.6 & 76.0 & 80.0$^\dagger$ & 74.0 & 73.5 & 74.71 \\
SFT + RAG (fixed)              & 68.0 & 84.0$^\dagger$ & 77.6 & 82.0$^\dagger$ & 82.0$^\dagger$ & 80.0$^\dagger$ & 81.6$^\dagger$ & 79.31 \\
SFT + RAG (structure-aware)    & 70.0 & 86.0$^\dagger$ & 69.4 & 82.0$^\dagger$ & 78.0 & 76.0 & 77.6 & 77.01 \\
RAFT (fixed)                   & 70.0 & 80.0$^\dagger$ & 79.6 & 78.0 & 88.0$^\dagger$ & 66.0 & 83.7$^\dagger$ & 77.87 \\
RAFT (structure-aware)         & 72.0 & 70.0 & 75.5 & 72.0 & 80.0$^\dagger$ & 72.0 & 75.5 & 73.85 \\
\bottomrule
\end{tabular*}

\end{table*}

\subsection{Difficulty Structure Across Examinations}\label{sec:results:difficulty}
 
Aggregation across examinations conceals substantial heterogeneity. Table~\ref{tab:full_matrix} reports every configuration on every administration, so that the dispersion underlying the pooled figures of Table~\ref{tbl:main_results} is visible in full.
 
The spread is wide. The highest single-paper score in the study is 90.0\% on the March 2021 PWR paper, and the lowest is 42.0\% on the March 2020 BWR paper, a range of 48 percentage points across the matrix. The highest score is not recorded by the configuration that passes the most examinations. It belongs to SFT with structure-aware chunking, whereas SFT with fixed-size chunking peaks at 88.0\% on the 2015 and 2021 PWR papers. At 50 items per paper the difference between these cells is one item and lies well inside the sampling interval of Section~\ref{sec:methodology:eval_protocol}. We record the observation to illustrate why the best single-paper score is not an admissible basis for selecting a configuration, and we rest no claim on it.
 
Counting passes by sitting rather than by configuration exposes a pronounced year effect. The matrix contains sixteen configuration-paper cells per sitting, and the fourteen papers together yield twenty-six passing cells. The March 2019 sitting accounts for nine of them, more than any other year. Every fine-tuned configuration meets the criterion on the 2019 PWR paper, at between 80.0\% and 84.0\%, and four of the five also meet it on the 2019 BWR paper, while no base configuration exceeds 74.0\% on either. The March 2021 sitting follows with seven. At the other extreme the March 2017 sitting is the only one on which no configuration meets the criterion on either paper. The best result there is 79.6\%, which corresponds to 39 of the 49 scored items and falls one item short of the 40 required. The 2015, 2018 and 2020 sittings yield two passing cells each.
 
The difficulty ordering is not uniform across configurations. Taking each configuration's weakest single paper, no sitting accounts for more than three of the eight. Three configurations are weakest in 2016, two in 2015, two in 2020 and one in 2017. Year difficulty is therefore partly dependent on model state rather than an intrinsic property of the paper, and a sitting that is hard for an off-the-shelf model need not be hard for its fine-tuned descendant.
 
The March 2015 BWR paper behaves differently again. The base model scores 48.0\% on it. That is only 3.7 points below its BWR average, so the paper is not intrinsically hard. Fine-tuned configurations fall far below their own averages on it. SFT falls by 10.7 points to 64.0\% and SFT with fixed-size retrieval falls by 11.3 points to 68.0\%. The deficit grows with fine-tuning rather than shrinking. This is the signature of a training-coverage gap rather than of item difficulty. The paper is also consequential. It accounts for the entire gap between BWR and PWR pooled accuracy. With it excluded, BWR accuracy under the best configuration would be 81.2\% against 80.23\% on PWR.

\subsection{Scoring Convention and Sensitivity}\label{sec:results:scoring}
The strict convention scores an unextractable response as incorrect and holds the denominator fixed. The alternative convention conditions accuracy on successful extraction and removes such items from the denominator. Under the alternative the effective denominator varies between conditions, which makes configurations non-comparable. Table~\ref{tbl:scoring} reports both.
\begin{table*}[width=\textwidth,cols=5,pos=h]
\caption{Combined accuracy in percent under the two scoring conventions, with the per-examination pass count. The extraction-conditioned figures use a denominator that varies between 695 and 697 across configurations.}\label{tbl:scoring}
\begin{tabular*}{\tblwidth}{@{\extracolsep{\fill}} Lcccc @{}}
\toprule
\textbf{Configuration} & \textbf{Extraction-cond.} & \textbf{Strict} & \textbf{Difference} & \textbf{Passes} \\
\midrule
Base, no retrieval           & 51.9 & 51.94 & $\pm0.0$ & 0/14 \\
Base + RAG (fixed)           & 62.4 & 62.27 & $-0.1$ & 0/14 \\
Base + RAG (structure-aware) & 64.8 & 64.56 & $-0.2$ & 0/14 \\
SFT, no retrieval            & 75.6 & 75.61 & $\pm0.0$ & 4/14 \\
\textbf{SFT + RAG (fixed)}   & 79.8 & 79.77 & $-0.0$ & \textbf{8/14} \\
SFT + RAG (structure-aware)  & 77.6 & 77.62 & $-0.0$ & 5/14 \\
RAFT (fixed)                 & 77.5 & 77.47 & $-0.0$ & 5/14 \\
RAFT (structure-aware)       & 75.5 & 75.47 & $+0.0$ & 4/14 \\
\bottomrule
\end{tabular*}
\end{table*}

The choice of convention does not affect the ordering of configurations, the per-examination pass count, or the position of the best pooled result relative to the threshold at the reported precision. The best configuration reaches 79.8\% under the extraction-conditioned convention and 79.77\% under the strict convention. The PWR figure of 80.23\% clears the mark under both. We therefore retain the strict convention because it preserves a common denominator and mirrors the examination rule that an unanswered item receives no credit.

The three excluded items require separate treatment. They are removed from all conditions from the primary analysis for the reason given in Section~\ref{sec:methodology:eval_protocol}. Treating all three as incorrect and restoring the original 700 administered slots changes the best configuration to 80.00\% on PWR, 78.86\% on BWR and 79.43\% pooled. The PWR figure remains at the criterion, which the NRC standard treats as passing since no rounding is required. No per-examination pass status changes under this treatment.

The width of the confidence intervals bounds what can be claimed. At 697 items the Wilson score 95\% interval for the best configuration is 79.77\% with bounds of 76.6\% and 82.6\%. This interval spans the threshold. The corresponding interval for the base model without retrieval is 51.94\% with bounds of 48.2\% and 55.6\%. The separation between fine-tuned and base configurations is far larger than the interval width. The position of the best configuration relative to the 80\% mark is not resolved at this sample size. Intervals among the four fine-tuned configurations overlap substantially. We therefore treat the chunking reversal and the RAFT deficit as robust in direction, on the strength of their consistency across reactor types and chunking strategies. We treat them as modest in magnitude. We claim no statistically significant separation among fine-tuned configurations from the pooled figures alone.

% =========================================================
% 4. DISCUSSION
% =========================================================
\section{Discussion}\label{sec:discussion}
This section interprets the pass count, the choice of evaluation unit, the two negative findings on retrieval design and the implications for deployment.

\subsection{What the Result Establishes}\label{sec:discussion:establishes}
An openly available 31-billion-parameter multimodal model fine-tuned with distilled CoT supervision and paired with fixed-size chunking retrieval satisfied the 80\% criterion on 8 of the 14 papers administered at the March sitting from 2015 to 2021, while no configuration without fine-tuning satisfied it on any. The appropriate reading is that the pipeline reaches the vicinity of the criterion rather than clearing it with margin. Pooled accuracy is 79.77\% under the strict convention, and the confidence interval spans the threshold. A human candidate who passed 8 of 14 sittings would not be described as reliably qualified.
 
Two readings must be kept apart. The claim that the pipeline passes the NRC examination is not supported, and we do not make it. The claim that is supported concerns a transformation: the gap between an off-the-shelf open model at 51.94\%, a score at which no human candidate would be licensed and which clears the criterion on none of the fourteen papers, and a configuration that meets the regulator's own criterion on a majority of them, is closable by domain adaptation alone. The practical significance of that transformation is that the closing techniques all run entirely on a single workstation. Data governance, auditability and on-premise deployment are frequently prerequisites in this sector, and the deployed artifact satisfies them: the merged adapter and the sparse index require no proprietary model, no external inference endpoint and no network access at run time. One external dependency exists at construction time, since the CoT rationales were distilled from a proprietary teacher, but it is confined to a single offline step and leaves no residual requirement in the deployed system.

\subsection{The Unit of Evaluation}\label{sec:discussion:unit}
We report per-examination outcomes because that is the unit at which the criterion operates. A candidate passes or fails a specific administration and never a weighted average across years. Pooled accuracy ranks PWR above BWR at 80.23\% against 79.31\%. It would support the statement that the model fails on BWR content. The per-examination count reverses this. The model met the criterion on 5 of 7 BWR papers against 3 of 7 PWR papers. The pooled ordering is produced almost entirely by one anomalous BWR paper. Where a benchmark inherits its criterion from an external standard we suggest that it should also inherit the unit of assessment of that standard. Reporting only pooled accuracy can invert the qualitative conclusion.

\subsection{Retrieval Design Depends on the Training State}\label{sec:discussion:chunking}
Structure-aware chunking is preferable before fine-tuning and fixed-size chunking after. This holds without exception in our pooled comparisons. A plausible mechanism follows from what the two schemes supply. Structure-aware chunks are section-level and self-contained. They carry a complete expository treatment of a topic. Fixed-size chunks are shorter and more numerous and overlap by 20\%. Any given assertion therefore recurs across several retrievable units.

The base model lacks an internal organization of the domain. It must reconstruct an explanation from the retrieved text and therefore benefits from narrative coherence. The fine-tuned model already possesses that organization. It needs only the specific quantity or relation on which the item turns. For that purpose the redundancy of overlapping windows raises the probability that the decisive fact appears somewhere in the context. The loss of narrative continuity costs little. The practical implication is that chunking cannot be tuned against an off-the-shelf model and then inherited by its fine-tuned descendant. That practice is nonetheless common.

\subsection{Why Retrieval-Conditioned Training Did Not Help}\label{sec:discussion:raft}
RAFT trails plain SFT by 2.3 and 2.2 points under the two chunking strategies and in both reactor types. The near-constancy of the deficit is inconsistent with sampling noise. It is also inconsistent with explanations that predict interaction with chunk granularity. We advance three candidate mechanisms in decreasing order of fit to the observed pattern.

The first concerns a mismatch internal to the RAFT training example identified in Section~\ref{sec:methodology:cot}. The context block shown to the student is sparse-retrieved. The rationale it is trained to reproduce was authored by the teacher against dense-retrieved passages, under an instruction to ground the explanation in those passages. The student is therefore trained on examples in which the target text cites evidence that may be absent from the accompanying context. Such supervision does not teach robustness to distracting retrieval. It instead associates confident assertion with contexts that do not license it. The predicted consequence is a uniform degradation, which is what we observe. The hypothesis is testable by measuring passage overlap between the teacher context and the student context across the training set. We identify that measurement as the primary follow-up to this study. We state it as a hypothesis because the pipeline was not designed to vary this factor and no experiment reported here isolates it.

The second concerns the retrieval corpus. Sparse lexical retrieval over an authoritative and tightly scoped handbook corpus returns relatively on-topic passages. RAFT is designed to confer robustness against distracting retrieval. Where retrieval is already clean, that robustness yields little benefit. Its cost in model capacity is unchanged. The third concerns capacity itself. At 3{,}577 items and rank 16 the capacity allocated to retrieval robustness may trade against capacity for the target reasoning. Under noisier or broader corpora the balance between the two paradigms may differ. We do not claim the ordering generalizes beyond this setting.

\subsection{Distinguishing Hard Examinations from Coverage Gaps}\label{sec:discussion:difficulty}
The March 2020 PWR paper and the March 2015 BWR paper both depart from the aggregate pattern, and they depart from it in different ways. The comparison that separates them is the deficit of a paper relative to a configuration's own average, evaluated before and after fine-tuning.
 
On the 2020 PWR paper both the base model and the strongest configuration score below their own PWR averages, by 4.2 and 6.2 points respectively. The paper is harder for the adapted model in roughly the measure that it is harder for the unadapted one. That is the signature of item difficulty. On the 2015 BWR paper the base model sits 3.7 points below its BWR average, which is unremarkable, while SFT falls 10.7 points and SFT with fixed-size retrieval falls 11.3 points below theirs. The deficit widens by a factor of three once fine-tuning is applied. Difficulty intrinsic to the items cannot produce that asymmetry, since it would depress the base model as well. What the pattern points to instead is content underrepresented in the bank-derived training corpus.
 
The diagnostic generalizes. Comparing the deficit of a paper under the base model against its deficit under the fine-tuned model separates item difficulty from training-coverage gaps using only quantities a benchmark already produces, with no additional experiments. It also carries an operational reading. A deployed assistant whose training corpus underrepresents a topic will not signal that fact through degraded confidence. Here it manifested as an accuracy drop of more than 11 points on one paper with no corresponding drop for the base model. Coverage auditing of the fine-tuning corpus is therefore required in a safety-critical deployment.

\subsection{Relation to Licensure Benchmarks in Other Professions}\label{sec:discussion:relation}
The reported outcome is weaker in absolute terms than the headline results on medical and legal licensure, where frontier models have cleared the USMLE across all three steps without domain-specific fine-tuning~\cite{nori2023gpt4usmle} and the Uniform Bar Examination near the 90th percentile of human candidates~\cite{katz2024gpt4bar}. Three differences make the comparison less unfavourable than the raw numbers suggest, and one makes it more so.
\par The first difference is the size of the model. Our subject has 31 billion parameters, is openly available and runs on one workstation, whereas the cited results were obtained from proprietary frontier systems accessed over a network. The second is the corpus. Clinical and legal text is abundantly represented in web-scale pretraining data, whereas reactor operations knowledge is concentrated in regulatory and vendor documents that are neither abundant nor uniformly indexed, so the base rate of 51.94\% is closer to chance-plus-partial-knowledge than the corresponding medical or legal base rates. The third is the criterion. A USMLE or bar result is reported as an aggregate percentile or a pass on a single administration, whereas we score fourteen separate administrations independently against a threshold that forbids rounding, which is the strictest of the three conventions and the one least forgiving of variance.
\par The difference that cuts the other way is consequence. An error on a bar examination item is an error on an examination. The domain in which a deployed reactor assistant would operate tolerates a far lower error rate than the 20\% residual we report, which is the reason Section~\ref{sec:discussion:deployment} confines the recommended role to an assistive one and Section~\ref{sec:discussion:difficulty} treats coverage auditing as mandatory rather than advisable. We therefore read our result as evidence about feasibility of adaptation and not as evidence of parity with the professional-licensure literature.

\subsection{Implications for Deployment}\label{sec:discussion:deployment}
These results indicate that fine-tuning and retrieval together can bring an open model near operator-level command of engineering fundamentals. They do not indicate readiness for autonomous action in a control room. The instrument used here measures generic technical knowledge under idealized conditions. It contains no questions on regulations, technical specifications or emergency operating procedures. It does not exercise real-time judgment, situational awareness or accountability.

The appropriate role at this competence level is assistive. Such a system can surface relevant reference material, cross-check reasoning and retrieve authoritative sources while a licensed operator retains decision authority. Roughly one item in five is answered incorrectly even in the best configuration. This indicates that human oversight is required. Any deployment must ground assertions in retrievable sources so that they can be traced and verified.

\subsection{Scope of the Claims and Threats to Validity}\label{sec:discussion:limitations}
The evaluation instrument, the passing criterion and the unit of assessment are all inherited from the regulator rather than selected by the authors, the evaluation set is a census of the March administrations of a closed program rather than a sample, and every configuration is scored on an identical item set under deterministic decoding. These properties remove several degrees of freedom that would otherwise bear on the credibility of the reported numbers. What follows bounds what the numbers can be taken to mean. The instrument covers generic fundamentals only, so the results say nothing about competence on plant-specific systems, regulations or procedures, which the remaining stages of licensing assess. Deduplication by normalized question text removes verbatim overlap but cannot remove the residual similarity between a derived item and its bank ancestor. By NRC construction 5 of 50 items per paper are such derivations~\cite{nrc_gfe_history}. A bounded optimism therefore remains. The evaluation covers the March sitting only. The pass count would differ under a different selection of administrations.

The deduplication procedure matches on normalized question text, and exact matching is defeated by formatting differences between the two sources. We state the resulting exposure in the terms least favourable to the study. The 711 removals correspond to 472 distinct normalized evaluation-question signatures. Because signatures recur across administrations and reactor types, those signatures occur in 573 of the 700 administered item rows, leaving 127 rows with no exact textual match in the banks. After excluding the three deleted item slots from primary scoring, 124 scored items remain unmatched. By NRC construction, 140 administered item slots should lack an exact bank counterpart~\cite{nrc_gfe_history}. However, this design-based count is not directly comparable with the 124 scored unmatched rows because repeated question signatures and formatting differences affect the item-level matching count. To determine the extent of potential residual overlap, we computed the near-duplicate structure that exact matching cannot see. Each of the 124 scored unmatched evaluation items was scored against every training item by cosine similarity over character 3-gram TF-IDF vectors, and the maximum over training items was retained. Of the 124, 62 exceed a similarity of 0.90, and 31 fall between 0.75 and 0.90. Manual inspection shows that these ranges contain both near-identical questions and substantively modified variants. Accordingly, 0.75 is used as a deliberately conservative sensitivity cutoff rather than as a verified classifier of contamination. Removing all 93 scored unmatched evaluation items above 0.75 leaves 604 items and changes the pooled accuracy of the best configuration from 79.77\% to 79.80\%, and changes the per-examination pass count from 8 of 14 to 9 of 14. The pooled accuracy of the best configuration is therefore essentially unchanged under this restriction. We report the unrestricted figures as primary because the restricted set is no longer the instrument the regulator administered, and we report the restricted figures as a deliberately conservative sensitivity analysis of what potential residual overlap could contribute. A further 121 training items carry fewer than four parsed answer options and 108 carry none, the option text having remained inside the question stem. The teacher received an empty option block for those items and the same items were used for fine-tuning. We study one model family and one sparse retriever. The efficacy of dense or hybrid retrieval and of other model scales is therefore unknown. The chunking reversal in particular may not survive a change of retriever. Individual examinations carry 50 items. The 95\% interval at that size is roughly 11 points in each direction. Per-examination outcomes are therefore informative only in aggregate. The closed four-choice format does not capture open-ended reasoning or confidence calibration. We do not quantify hallucination rates or the faithfulness of generated rationales. An item answered correctly with faulty reasoning is scored identically to one answered correctly with sound reasoning. The CoT targets derive from a single teacher. Systematic reasoning biases in that teacher may therefore propagate to the student undetected. Each fine-tuned configuration is the product of a single training run at one random seed. Decoding is deterministic, so the reported accuracies are exactly reproducible from a given adapter, but the run-to-run variance induced by initialization and data order is not characterized, and it is the component of variance most likely to account for differences of two to three points among the fine-tuned configurations. For the same reason we report no paired item-level significance test on the between-configuration contrasts. The natural instrument is McNemar's test on the discordant items of each pair, computed under a correction for the twenty-eight pairwise comparisons the design admits, and we identify it together with a seed-replication study as the statistical work that would convert the direction-consistent findings of Sections~\ref{sec:results:chunking} and~\ref{sec:results:raft} into effect estimates with calibrated uncertainty. Finally, the evaluation is closed-book with respect to reasoning quality in a further sense: the pass criterion rewards the selected option and is indifferent to whether the retrieved context was used, so we cannot distinguish an item answered from the retrieved passage from one answered from parametric memory while the passage was ignored.

% =========================================================
% 5. CONCLUSION
% =========================================================
\section{Conclusion}\label{sec:conclusion}
An open-weight 31-billion-parameter multimodal model, benchmarked across eight configurations on every NRC Generic Fundamentals Examination administered at the March sitting from 2015 to 2021 and scored paper by paper against the same 80\% criterion required of human candidates, satisfies that criterion on 8 of the 14 papers when fine-tuned on distilled CoT rationales and paired with fixed-size chunking BM25 retrieval over the DOE Fundamentals Handbook corpus. No configuration without fine-tuning satisfies it on any paper, and the two groups do not overlap on this metric. The pipeline that produces this outcome, from low-rank fine-tuning through adapter merging and MLX inference, executes end to end on a single workstation. Pooled accuracy of 79.77\% sits below the threshold with a confidence interval that spans it. The PWR figure of 80.23\% sits above it. The result is therefore properly read as approaching operator-level command of engineering fundamentals rather than reliably achieving it.
 
The central contribution is a transformation rather than a measurement. The same open-weight model answers 51.94\% of the items correctly out of the box, a score at which no human candidate would be licensed, and satisfies the criterion on none of the fourteen papers. After domain adaptation it reaches 80.23\% on the PWR item set and meets the criterion on 8 of the 14 papers. The distance between an off-the-shelf model that fails the assessment outright and one that meets a regulator's own criterion on a majority of administrations is therefore closable by fine-tuning and retrieval alone, without recourse to a larger model or to a different architecture. The criterion is not one we chose. It is the threshold the NRC applies to every human candidate, and the papers are the instruments actually administered, so the comparison is made on terms the regulator already set rather than on a benchmark constructed for the purpose.
 
The second contribution concerns what the result costs to obtain. Fine-tuning, adapter merging, format conversion and inference all execute on a single commodity workstation with unified memory, at a parameter count small enough that the weights, the training corpus and the retrieval index can be held and audited in one place. Retrieval is served by a sparse index over seven publicly available DOE handbooks, so the deployed system depends on no proprietary model, no external inference endpoint and no network access at run time. One external dependency does exist and we state it plainly. The CoT rationales were distilled from a proprietary teacher model. That dependency is confined to a single offline step in training-data construction, and the resulting adapter and index carry no residual requirement for it, so the artifact that would be deployed is self-contained even though the pipeline that produced it was not. For a sector in which data governance, auditability and on-premise operation are frequently prerequisites rather than preferences, this property is a substantive part of the result rather than an implementation detail.
 
Three further findings refine how such systems should be built. First, the optimal chunking strategy reverses with the training state of the model. Structure-aware chunking favors the base model and fixed-size sliding windows favor the fine-tuned model. Retrieval components must therefore be co-designed with the fine-tuning strategy rather than selected in advance. Second, RAFT underperforms plain SFT under matched retrieval by a margin that is small but consistent across chunking strategies and reactor types. We attribute this provisionally to a mismatch between the retriever that supplied the evidence of the teacher and the retriever that supplies the context of the student. A passage-overlap measurement would settle the question. Third, comparing the deficit of a paper under the base model against its deficit after fine-tuning separates intrinsic item difficulty from gaps in training coverage. The diagnostic requires no additional experiments. It surfaced a coverage gap invisible to pooled accuracy.
 
These results position a fine-tuned and retrieval-grounded language model as an assistive tool that augments rather than replaces the judgment of a licensed operator. Roughly one item in five is still answered incorrectly in the best configuration, and the instrument covers generic fundamentals only. Future work will measure teacher and student context overlap to test the RAFT hypothesis directly. It will also route chunk-attached figure rasters through the multimodal interface at inference, which Section~\ref{sec:results:multimodal} identifies as the retrieval pathway the present pipeline leaves unused, and compare dense and hybrid retrieval against the sparse baseline established here. It will also seek an explanation for the reactor-type asymmetry in retrieval gain reported in Section~\ref{sec:results:asymmetry}, which a decomposition by topic rather than by reactor type would localize. Establishing trust in safety-critical settings requires evaluating not only whether answers are correct but whether the reasoning supporting them is faithful. We identify calibration measurement and faithfulness metrics such as entailment checking and citation accuracy as the necessary next step.
 
\section*{Data availability}
The evaluation items, the question banks and the retrieval corpus are drawn entirely from publicly available NRC and DOE publications cited in Section~\ref{sec:methodology:dataset}. The prompt templates, chunking parameters and evaluation scripts needed to reproduce the reported results are available from the corresponding author on reasonable request.
 
\section*{Declaration of competing interest}
The authors declare that they have no known competing financial interests or personal relationships that could have appeared to influence the work reported in this paper.
 
\section*{Acknowledgements}
Corresponding author S.B.A. acknowledges support from the U.S. Nuclear Regulatory Commission under grant numbers 31310025M0012 and 31310024M0041.
 
\printcredits
\bibliographystyle{model1-num-names}
\bibliography{cas-refs}
\end{document}